\documentclass{article}

\usepackage{arxiv}

\usepackage[utf8]{inputenc} 
\usepackage[T1]{fontenc}    
\usepackage{url}            
\usepackage{booktabs}       
\usepackage{amsfonts}       
\usepackage{nicefrac}       
\usepackage{microtype}      
\usepackage{lipsum}		
\usepackage{graphicx}
\usepackage{doi}
\usepackage{amsmath}
\usepackage{algorithm}
\usepackage[noend]{algpseudocode}
\usepackage{multirow}
\usepackage{wrapfig}
\usepackage{svg}
\usepackage{tabularx}
\usepackage{bm}
\usepackage{multicol}
\usepackage{caption}

\usepackage{caption}

\usepackage[numbers]{natbib} 
\usepackage{hyperref}        
\hypersetup{hidelinks}

\title{Offline reinforcement learning for job-shop scheduling problems}

\author{Imanol Echeverria\thanks{Corresponding author.} \\
    TECNALIA, Basque Research and Technology Alliance (BRTA) \\
	\texttt{imanol.echeverria@tecnalia.com} \\
	\And
	Maialen Murua\\
	TECNALIA, Basque Research and Technology Alliance (BRTA)\\
	\texttt{maialen.murua@tecnalia.com} \\
 	\And
	Roberto Santana\\
	Computer Science and Artificial Intelligence Department, University of the Basque Country\\
	\texttt{roberto.santana@ehu.eus} \\
}

\date{}

\renewcommand{\headeright}{}
\renewcommand{\undertitle}{}
\renewcommand{\shorttitle}{}

\hypersetup{
pdftitle={Offline reinforcement learning for job-shop scheduling problems},
pdfsubject={cs.IA},
pdfauthor={Imanol Echeverria},
pdfkeywords={Offline reinforcement learning for job-shop scheduling problems},
}

\begin{document}
\maketitle

\begin{abstract}
Recent advances in deep learning have shown significant potential for solving combinatorial optimization problems in real-time. Unlike traditional methods, deep learning can generate high-quality solutions efficiently, which is crucial for applications like routing and scheduling. However, existing approaches like deep reinforcement learning (RL) and behavioral cloning have notable limitations, with deep RL suffering from slow learning and behavioral cloning relying solely on expert actions, which can lead to generalization issues and neglect of the optimization objective. This paper introduces a novel offline RL method designed for combinatorial optimization problems with complex constraints, where the state is represented as a heterogeneous graph and the action space is variable. Our approach encodes actions in edge attributes and balances expected rewards with the imitation of expert solutions. We demonstrate the effectiveness of this method on job-shop scheduling and flexible job-shop scheduling benchmarks, achieving superior performance compared to state-of-the-art techniques.
\end{abstract}

\begin{multicols*}{2}
\raggedcolumns

\section{Introduction}\label{sec:introduction}
Recently, the use of deep learning has emerged as a relevant field of study for solving combinatorial optimization problems \cite{khalil2017learning}. One of the main advantages of deep learning over previous methods is its ability to generate high-quality solutions in real-time, unlike exact or metaheuristic-based approaches whose execution times increase with the complexity of the instances to be solved \cite{oliveto2007time}. Real-time resolution offers significant benefits in multiple domains, such as routing or scheduling \cite{paschos2014applications}, by enabling rapid responses to disruptive events and facilitating resource optimization through efficient scenario simulation.

These methods have largely relied on generating policies through deep reinforcement learning (DRL) \cite{wang2021deep}. This involves training an agent to learn a policy by interacting with an environment and receiving feedback in the form of rewards or penalties based on its actions. Another strategy, which falls under the umbrella of behavioral cloning (BC) methods, involves generating a set of optimal solutions and training the policy to mimic the generation of these solutions \cite{bengio2021machine}.

Nevertheless, both learning methods have significant shortcomings. DRL-based approaches, which begin by exploring the solution space randomly, can be slow to learn and may not always succeed in finding optimal solutions, especially in complex scenarios. On the other hand, BC-based methods do not account for rewards and hence ignore the optimization objective, relying solely on expert actions. Furthermore, this reliance on expert observations can lead to generalization issues when the policy encounters new situations or suboptimal states due to flaws in the policy \cite{ross2010efficient}.

Offline RL has been introduced as a hybrid approach that leverages the strengths of both methods while requiring only a set of example instances for training  \cite{levine2020offline}. This method reduces dependency on real-time data and allows the policy to learn from a large, diverse set of historical data. Although its application to combinatorial optimization problems is still under-explored \cite{bengio2021machine}, offline DRL is increasingly recognized as a preferable method over BC, even when optimal solutions are available \cite{kumar2022should}. Therefore, this paper seeks to explore this promising area by proposing a new offline RL algorithm specifically tailored for combinatorial optimization challenges that involve multiple difficult constraints and requires real-time solutions. 

Graphs are increasingly used to represent problem instances with complex dependencies among their components \cite{khalil2017learning}. They can contain additional problem-specific information through data in nodes and edges, which connect in various ways to show different relationships. Heterogeneous graphs, in particular, have different types of nodes and edges, each with unique attributes and roles, adding layers of complexity to the representation. This complexity is necessary to accurately capture the many aspects of real-world problems, allowing for more precise and effective solutions. However, using offline DRL algorithms in such graph-based problems poses substantial challenges because most DRL approaches assume a fixed state and action space, typically represented as vectors or matrices, allowing for straightforward concatenation during model training. Despite these challenges, there is significant potential for DRL algorithms in real-world applications using graphs, such as recommendation systems \cite{shi2018heterogeneous} or navigation solutions \cite{zhou2023learning}, to effectively use the rich information in heterogeneous graphs \cite{munikoti2023challenges}.

The main goal of this paper is to propose more efficient methods for solving combinatorial optimization problems that incorporate difficult constraints and require real-time solutions. For this purpose, we introduce a new offline RL method that considers problems where the state space is represented as a heterogeneous graph and the action space is variable. Scheduling problems, such as the job-shop scheduling problem (JSSP) and the flexible JSSP (FJSSP), usually have more constraints than routing problems due to the need to account for the sequential ordering of operations, machine availability, and processing times. As a case study, we have used these two scheduling problems to demonstrate the effectiveness of our approach. The contributions of this paper are summarized as follows:

\begin{itemize}
\item We model the JSSP and FJSSP as a Markov Decision Process (MDPs) with limited visible operations and simultaneous assignment of multiple tasks, reducing state transitions and model evaluations.
\item Introducing a new offline RL action-value based method for problems where the state is represented as a heterogeneous graph and the action space is variable. In our method, we utilize edge attributes to encode which action has been taken at each step.
\item Proposing a new loss function for an offline DRL approach that balances expected rewards with a loss term based on a classification metric related to the ability to imitate expert solutions.
\item The proposed method has been tested against five well-known benchmarks (one for the JSSP and five benchmarks for the FJSSP), outperforming various state-of-the-art methods for each problem.
\end{itemize}

The organization of the paper is as follows: In Section \ref{sec:relatedWork}, we review recent advancements in deep learning approaches for combinatorial optimization, particularly focusing on scheduling problems. Section \ref{sec:preliminaries} details the formulation of the JSSP and FJSSP, emphasizing their complexity and constraints. Our proposed offline RL method, tailored for graph-based representations with variable action spaces, is thoroughly described in Section \ref{sec:proposedmethod}. Section \ref{sec:experimentalresult} presents the experimental setup and results, demonstrating the effectiveness of our approach on several benchmarks. Finally, Section \ref{sec:conclusion} summarizes our findings and suggests potential directions for future research in this area.

\section{Related work}\label{sec:relatedWork}
To organize the approaches that address scheduling problems in real-time, we will first explore RL-based techniques, beginning with the methods that solve the JSSP and then moving on to the FJSSP. Subsequently, methods using other approaches, such as BC or self-supervised learning, will be examined.

\begin{table*}[!t]
\footnotesize
\caption{Survey of DL-methods for solving scheduling problems in real-time.}
\label{tab:related}
\begin{tabularx}{\textwidth}{l X X X X X}
        \hline
        Method & Problem & Learning method & Algorithm &  Type & Year\\
        \hline
        \hline
        \cite{zhang2020learning}&JSSP&RL&PPO& Constructive&2020\\
        \cite{park2021schedulenet}&JSSP&RL&REINFORCE& Constructive&2021\\
        \cite{chen2022deep}&JSSP&RL&Policy gradient & Constructive&2022\\
        \cite{tassel2023end}&JSSP&RL&Policy gradient & Constructive&2023\\
        \cite{zhang2024deep}&JSSP&RL&REINFORCE &Improvement&2024\\
        \cite{ho2024residual}&JSSP, FJSSP&RL&REINFORCE &Constructive&2024\\
        \cite{song2022flexible}&FJSSP&RL&PPO &Constructive&2022\\
        \cite{wang2023flexible}&FJSSP&RL&PPO&Constructive&2023\\
        \cite{echeverria2025diverse}&FJSSP&RL&PPO &Constructive&2024\\
        \cite{yuan2024solving} & FJSSP & RL & PPO & Constructive & 2024 \\
        \cite{huang2024learning} & FJSSP & RL & PPO & Constructive & 2024 \\
        \hline
        \cite{ingimundardottir2018discovering}&JSSP&BC&- &Constructive&2018\\
        \cite{li2024learning}&FSS&BC&- &Constructive&2022\\
        \cite{echeverria2025leveraging}&FJSSP&BC&-&Constructive&2024\\
        \cite{echeverria2024multi}&JSSP&BC&-&Constructive&2024\\
        \cite{corsini2024self}&JSSP&Self-supervised&-&Constructive&2024\\
        \hline
        \textbf{Ours}&JSSP, FJSSP&Offline RL&TD3 + BC&Constructive&2024\\
        \hline
    \end{tabularx}
\end{table*}

Methods for solving real-time scheduling problems often build upon previous research that addressed simpler combinatorial challenges as they contain less constraints, such as the Travelling Salesman Problem and the Capacitated Vehicle Routing Problem \cite{vinyals2015pointer, kwon2020pomo, kool2018attention}. Recent approaches \cite{kwon2020pomo} make use of transformer architectures \cite{vaswani2017attention}, modeling the problem as a sequence of elements where each element is defined by a distinct set of features. However, these methods often do not explicitly consider the different types of nodes or the attributes of the edges that connect them. 

In scheduling problems, which involve different types of entities (operations, jobs, and machines), most approaches model the problem as a graph since this facilitates effective problem modeling, albeit coupled with the use of specific neural networks that can process this type of representation. The most common way to generate solutions is constructively, where solutions are constructed iteratively: at each step, an element is selected based on its characteristics. For instance, in job scheduling, the process could be visualized as sequentially assigning operations to machines.

Employing this approach, in \cite{zhang2020learning}, the JSSP was modeled as a disjunctive graph, employing a graph neural network (GNN) to extract information from instances and the Proximal Policy Optimization (PPO) algorithm \cite{schulman2017proximal} to optimize its policy. Building on this framework, several methods have been proposed for the JSSP that vary in how the network is trained, the reward function used, or how the problem itself is modeled \cite{park2021schedulenet,chen2022deep, tassel2023end}. These methods leverage policy gradient algorithms, such as REINFORCE \cite{williams1992simple}, which focus on optimizing policies by directly estimating the gradient of the expected reward. While policy gradient methods, including PPO, are commonly used in reinforcement learning, their application in offline RL remains limited, as offline approaches typically favor techniques better suited for working with static datasets.

A different approach to the JSSP involves neural improvement methods. Unlike constructive methods, where solutions are built through the sequential assignment of operations to machines, these methods aim to enhance a solution by modifying the execution sequence of operations \cite{zhang2024deep}. These methods also utilize variants of the REINFORCE algorithm to generate the policy.

A notable approach for solving the FJSSP is detailed in \cite{song2022flexible}, where the problem is represented as a heterogeneous graph and optimized using PPO. A heterogeneous graph represents different types of entities (e.g., jobs and machines) and the relationships between them. Similar approaches to solve the FJSSP also utilize PPO with variations in how the state and action space are represented and how the policy is generated \cite{wang2023flexible, echeverria2025diverse, yuan2024solving}. Recently, \cite{huang2024learning} introduced the Graph Gated Channel Transformation with PPO to effectively handle scale changes and enhance policy performance across datasets.

Imitation learning serves as a less common but effective alternative to RL for policy generation. In this method, optimal solutions are generated with an optimization solver such as CPLEX or Gurobi, and the model is then trained to imitate how these solutions are constructed using BC. This approach enhances neural network performance due to the higher quality of the training set. Unlike RL, which requires exploring sub-optimal solutions—a time-consuming process—imitation learning directly leverages optimal solutions, streamlining the learning phase. Research presented in \cite{ingimundardottir2018discovering} developed a method using a mixed-integer programming solver to create optimal dispatching rules for the JSSP. Authors of \cite{li2024learning} used GNNs to imitate optimal sequences for the Flow Shop Scheduling (FSS), a simpler JSSP variant, showing improvements over RL-based neural networks. dditionally, \cite{echeverria2025leveraging} addressed the FJSSP by training a heterogeneous GNN in a supervised manner, achieving superior results compared to several DRL-based approaches. Moreover, \cite{echeverria2024multi} proposed a method for the JSSP, also based on BC, which allows multiple assignments to be made simultaneously. These types of approaches lack direct consideration of reward optimization, which can impair the policy's performance, as suggested in \cite{kumar2022should}.


Among the methods that neither use RL nor BC, the work of \cite{corsini2024self} introduces a self-supervised training strategy for the JSSP. This approach generates multiple solutions and uses the best one according to the problem's objectives as a pseudo-label. However, it should be noted that this method does not utilize rewards or expert solutions, which might affect both the speed and the quality of the learning process.

In Table \ref{tab:related}, a summary of the mentioned approaches is presented, focusing on which problem is solved, the learning method used, and the algorithms employed. As it can be observed, most approaches primarily utilize RL and policy gradient algorithms. A smaller number of contributions use BC, and one single approach uses a different method. This limitation in the variety of learning strategies for policy development suggests that the potential to find optimal policies is not being fully exploited. It is also noted that the majority of approaches propose solutions that are only tested on a single class of optimization problems.

\section{Preliminaries}\label{sec:preliminaries}

\subsection{Problem formulation}

An instance of the FJSSP problem is defined by a set of jobs $\mathcal{J} = \{j_1 , j_2, \ldots, j_n \}$, where each $j_i \in \mathcal{J}$ is composed of a set of operations $\mathcal{O}_{j_i} = \{ o_{i1}, o_{i2}, \ldots o_{im}\}$, and each operation can be performed on one or more machines from the set $\mathcal{M} = \{m_1, m_2, \ldots , m_p\}$. The JSSP is a specific case of this problem where operations can only be executed on a single machine. The processing time of operation $o_{ij}$ on machine $m_{k}$ is defined as $p_{ijk} \in \mathbb{R}^+$. We define $\mathcal{M}_{o_{ij}} \subseteq \mathcal{M}$ as the subset of machines on which that operation $o_{ij}$ can be processed, $\mathcal{O}{j_i} \subseteq \mathcal{O}$ as the set of operations that belong to the job $j_i$ where $\bigcup\limits_{i = 1}^n \mathcal{O}{j_i} = \mathcal{O}$, and $\mathcal{O}{m_k} \subseteq \mathcal{O}$ as the set of operations that can be performed on machine $m_k$. The execution of operations on machines must satisfy a series of constraints:

\begin{itemize}
\item All machines and jobs are available at time zero.
\item A machine can only execute one operation at a time, and the execution cannot be interrupted.
\item An operation can only be performed on one machine at a time.
\item The execution order of the set of operations $\mathcal{O}_{j_i}$ for every $j_i \in \mathcal{J}$ must be respected.
\item Job executions are independent of each other, meaning no operation from any job precedes or has priority over the operation of another job.
\end{itemize}

\begin{figure*}[!t]
  \centering
  \begin{minipage}[b]{0.49\textwidth}
    \includegraphics[width=\textwidth]{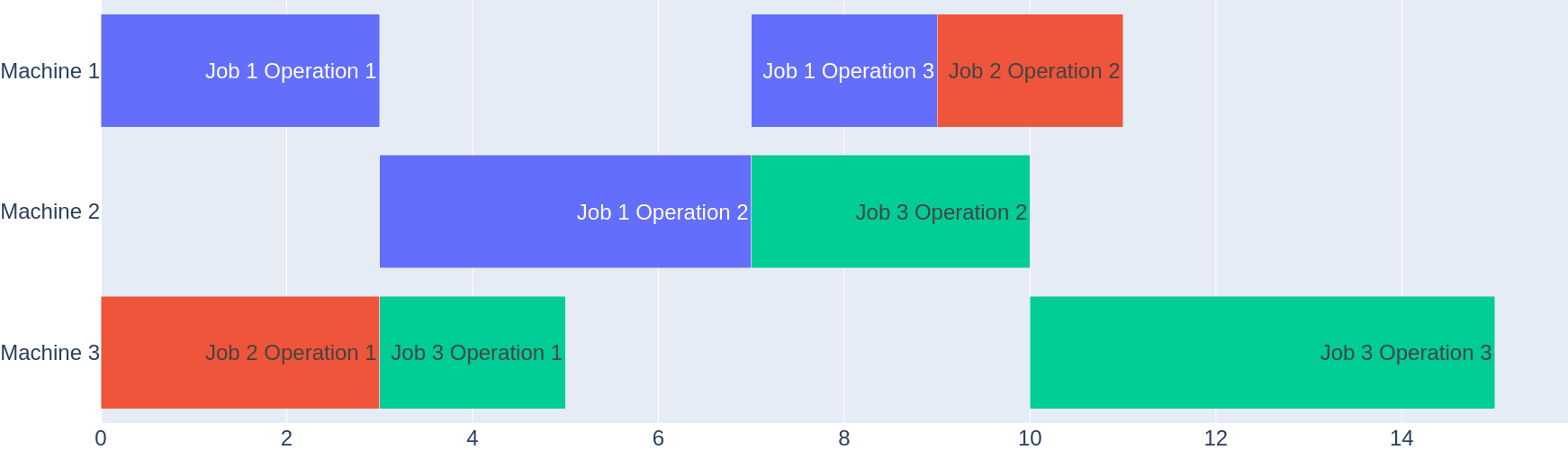}
    \caption{A solution to the JSSP instance with a makespan of 15.}
    \label{fig:sol}
  \end{minipage}
  \hfill
  \begin{minipage}[b]{0.49\textwidth}
    \includegraphics[width=\textwidth]{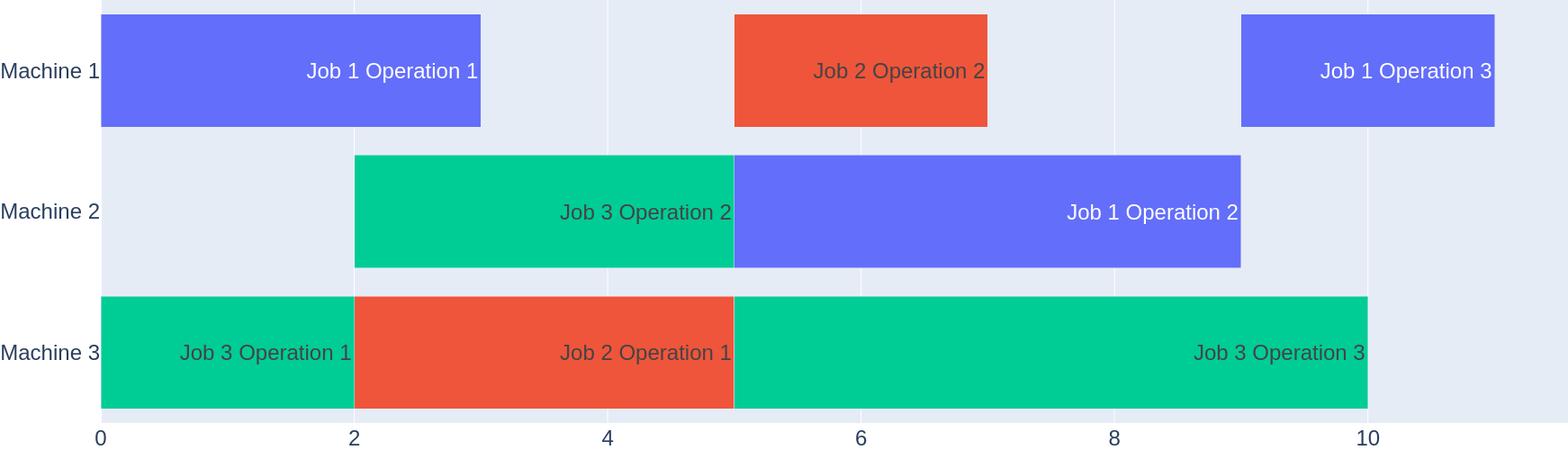}
    \caption{An optimal solution with a makespan of 11.}
    \label{fig:optsol}
  \end{minipage}
\end{figure*}

In essence, the FJSSP combines two problems: a machine selection problem, where the most suitable machine is chosen for each operation, a routing problem, and a sequencing or scheduling problem, where the sequence of operations on a machine needs to be determined. Given an assignment of operations to machines, the completion time of a job, $j_i$, is defined as $C_{j_i}$, and the makespan of a schedule is defined as $C_{max} = \max\limits_{j_i \in \mathcal{J}} C_{j_i}$, which is the most common objective to minimize.

Table \ref{tab:jssp_example} shows a small FJSSP instance with three machines and three jobs, each having three operations. The table lists which machines can perform each operation and their processing times. Figures \ref{fig:sol} and \ref{fig:optsol} illustrate two solutions with makespans of 15 and 11, respectively. The solution with a makespan of 11 is the optimal solution for this instance. Compared to the solution with a makespan of 15, the operations are distributed in a way that reduces machine idle times, thereby minimizing the makespan. The x-axis represents time, and the y-axis represents machine assignments, with colored bars indicating job operations (same color for the same job).

\begin{table}[H]
\caption{JSSP instance with 3 jobs, 3 machines, and 8 operations.}
\label{tab:jssp_example}
\begin{tabularx}{\columnwidth}{X X X X X X}
\toprule
Jobs  & Operations & \(m_1\) & \(m_2\) & \(m_3\) \\ \midrule
\(j_1\) & \(o_{11}\) & 3 & - & - \\
       & \(o_{12}\) & - & 4 & - \\
       & \(o_{13}\) & 2 & - & - \\
\hline
\(j_2\) & \(o_{21}\) & - & - & 3 \\
       & \(o_{22}\) & 2 & - & - \\
\hline
\(j_3\) & \(o_{31}\) & - & - & 2 \\
       & \(o_{32}\) & - & 3 & - \\
       & \(o_{33}\) & - & - & 5 \\
       \bottomrule
\end{tabularx}
\end{table}

\subsection{Offline RL and Behavioral Cloning}

RL is a method for solving tasks that are organized in sequences, known as a MDP, defined by the elements \(S\) (states), \(A\) (actions), \(R\) (rewards), \(p\) (transition probabilities), and \(\gamma\) (discount factor) \cite{sutton2018reinforcement}. In RL, an agent follows a policy \(\pi\), which can either directly map states to actions or assign probabilities to actions. The agent's objective is to maximize the expected total reward over time, represented as \(E_{\pi} \left[\sum_{t=0}^{\infty} \gamma^t r_{t+1}\right]\), where \(\gamma\) is the discount factor. This objective is evaluated using the value function \(Q_{\pi} (s, a) = E_{\pi} \left[\sum_{t=0}^{\infty} \gamma^t r_{t+1} \mid s_0 = s, a_0 = a\right]\), which estimates the expected rewards starting from state \(s\) and action \(a\).

In offline RL, there is a dataset \(\mathcal{D}\) that contains tuples of states, actions, and rewards. This dataset is used to train the policy without further interaction with the environment, addressing the challenges of direct interaction. By leveraging offline data, offline RL avoids the risk and expense associated with deploying exploratory policies in real-world settings. The dataset \(\mathcal{D}\) allows the agent to learn from a wide variety of experiences, including rare or unsafe states that might be difficult to encounter through online exploration.

BC is another approach that trains a policy by imitating an expert's actions. This type of imitation learning uses supervised learning to teach the policy to replicate actions from a dataset. The effectiveness of this method largely depends on the quality of the dataset used for training. While BC can be straightforward, it does not account for future rewards and may struggle with generalization to new situations.

\section{Method} \label{sec:proposedmethod}
In this section, we present our novel offline RL algorithm designed for combinatorial optimization problems with heterogeneous graph representations and variable action spaces. First, we model the JSSP and FJSSP as MDPs, capturing the complex dependencies between jobs and machines through a graph-based state representation. Additionally, we introduce a method for generating diverse experiences to enhance the policy’s ability to solve these problems efficiently in real-time.

\subsection{The JSSP and FJSSP as an MDP}

Before introducing our offline RL method, we describe how the JSSP and FJSSP have been modeled as an MDP. The MDP is structured through the definition of the state and action spaces, reward function, and transition function as follows:

\textbf{State space.} The state space is modeled using heterogeneous graphs, as described in \cite{echeverria2024multi}. At each timestep $t$, the state $s_t$ is represented by a heterogeneous graph $\mathcal{G}_t = (\mathcal{V}_t, \mathcal{E}_t)$, consisting of nodes for operations, jobs, and machines, along with six types of edges, both directed and undirected. To streamline decision-making, the number of visible operations per job is limited to the first operations of a job. This selective visibility is particularly beneficial for large instances, where operations with many pending tasks (i.e., the last operations) are less impactful for current assignments. Detailed descriptions of node and edge features are available in the Appendix \ref{appendix:nodefatures}.

\textbf{Action space.} The action space $\mathcal{A}_t$ at each timestep $t$ consists of feasible job-machine pairs. When a job is selected, its first unscheduled operation is chosen. To prevent an excessive number of choices, the action space is constrained by defining $t_e$ as the earliest time a machine can start a new operation and masking actions where the start time exceeds $t_e \times p$, where $p$ is a parameter slightly greater than one.

\textbf{Transition function.} The solution is constructed incrementally by assigning operations to machines. At each step, the policy can make multiple assignments, but with specific constraints: only one operation per job or one operation per machine can be assigned. In other words, it is not allowed to assign multiple operations from the same job; only the first available operation can be scheduled. Similarly, multiple operations cannot be assigned to a single machine simultaneously.

Once an operation is assigned, it is removed from the graph, and the edges of the corresponding job are updated to reflect the next operation to be processed. Additionally, the features of the remaining nodes are updated, and a new operation is added to the graph if there are pending tasks. The reason for allowing multiple assignments at once is to reduce the number of times the model is used to generate a solution, as repeatedly calling the model can become problematic in larger instances,  especially when real-time performance is required.

\textbf{Reward function.} The goal of the agent is to minimize the completion time of all operations, or makespan, similar to most approaches \cite{wang2023flexible, song2022flexible}. As suggested in previous works \cite{zhang2020learning}, one effective approach is to define the reward at timestep $t$ as $r(s_t, a_t, s_{t+1}) = C(s_t) - C(s_{t+1})$, where $C(s_t)$ represents the makespan at that step. By setting the discount factor $\gamma$ to 1, the cumulative reward becomes $\sum_{t=0}^{|\mathcal{O}|} r(s_t, a_t, s_{t+1}) = C(s_0) - C(s_{|\mathcal{O}|})$. If the initial makespan is zero or remains constant, maximizing this cumulative reward is equivalent to minimizing the final makespan $C_{s_{|\mathcal{O}|}}$ at the last state.

\subsection{Offline Reinforcement Learning for Heterogeneous Graphs}

Our offline RL approach for heterogeneous graphs, which we refer to as H-ORL, follows a minimalist strategy in offline RL, inspired by the work of \cite{fujimoto2021minimalist} and building upon the work proposed in \cite{echeverria2024multi}. The goal of our approach is to balance the maximization of expected rewards with the minimization of the discrepancy between expert actions and the policy's actions. To achieve this, we model the problem as a multi-objective optimization task, where the policy must simultaneously optimize for high rewards while ensuring that its actions remain close to those in the dataset.

We build our proposal on the established Twin Delayed Deep Deterministic Policy Gradient (TD3) algorithm \cite{fujimoto2018addressing}. In the standard TD3 framework, the policy \( \pi_{\theta} \), parameterized by $\theta$, is optimized by maximizing the expected Q-value, as represented by the equation:

\begin{equation}\pi_{\theta} = \text{argmax}_{\pi} \mathbb{E}_{(s,a) \sim \mathcal{D}} [Q(s, a ))] \end{equation}

However, this approach solely focuses on maximizing expected rewards, which may not be sufficient in offline settings where the policy must generalize well to unseen states while staying within the distribution of the training data. In our case, the dataset consists of optimal observations, so we want the policy to remain close to this type of action and avoid deviations.

To address this, we propose adding a term that measures this divergence in actions using the KL divergence loss. The KL divergence provides a more stable and interpretable measure of how closely the policy's action distribution aligns with the expert data, as supported by the findings in \cite{rusu2015policy}. The revised objective function is:
\begin{align}
\pi _{\theta} = \text{argmax}_{\pi} \mathbb{E}_{(s,a) \sim \mathcal{D}} & \Bigg[\lambda_{RL} Q(s, a) - \\ \nonumber
& \lambda_{BC} \; \pi_{\mathcal{D}}(a | s) \log \left({\frac {\pi_{\mathcal{D}}(a | s)}{\pi(a | s)}}\right) \Bigg] 
\end{align}
where $\lambda_{RL}$ and $\lambda_{BC}$ are adjustable parameters that control the influence of the reward maximization and behavior cloning terms, respectively, and $\pi(a | s)$ represents the probability distribution of actions in state $s$ generated by the policy, while $\pi_{\mathcal{D}}(a | s)$ denotes the optimal distribution. By fine-tuning these parameters, we can achieve a balanced approach that optimizes both the policy's performance and its adherence to expert behavior. In a manner similar to multi-objective optimization, the use of separate parameters facilitates a controlled trade-off between reward maximization and expert behavior alignment, providing flexibility in balancing these objectives during the learning process.

One of the significant challenges in applying this algorithm to combinatorial optimization problems modeled as graphs, such as those encountered in scheduling or routing, lies in the computation of the Q-value $Q(s, a)$. Unlike traditional environments where state and action spaces are fixed, graphs present a variable action space, making it difficult to apply standard neural network architectures directly.

\begin{figure*}[t]
    \centering
    \includegraphics[width=\textwidth]{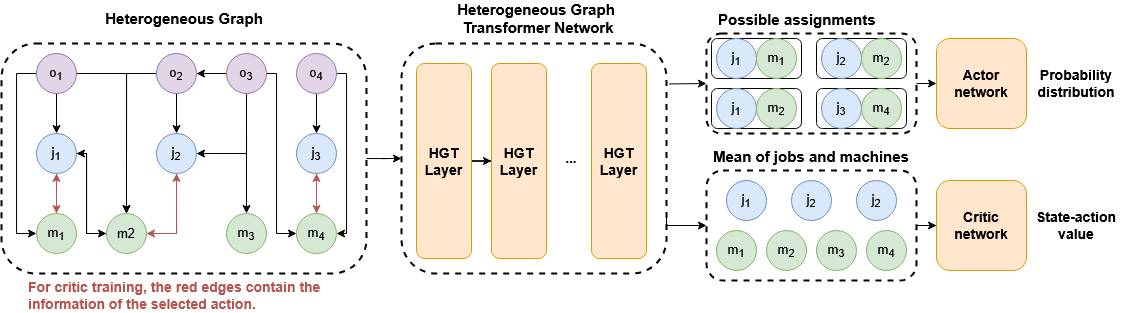}
    \caption{The model architecture of our approach.}
    \label{fig:model_arch}
\end{figure*}

To overcome this, we propose integrating the action information as an edge attribute within the graph structure. Specifically, in our scenario, where the nodes represent operations and machines in a scheduling problem, we already have edges linking these nodes with relevant attributes. By concatenating the action-related information with these existing attributes, we can preserve the flexibility of the graph representation while ensuring that the policy can effectively learn and apply the Q-value function.

Furthermore, given the variable nature of the action space, the use of the logarithm of the softmax function in the loss computation helps in differentiating and stabilizing the values, which is particularly important in a graph neural network context where maintaining numerical stability is crucial.

\subsection{Actor-critic architecture}

We utilize a Heterogeneous GNN architecture to extract features from the states, inspired by prior work using attention mechanisms \cite{hu2020heterogeneous}, which is referred to as the Heterogeneous Graph Transformer (HGT). Figure~\ref{fig:model_arch} shows the overall architecture of our approach.

As an example, we will show how the embedding of a job node \(j_i\) is calculated. This embedding is computed by aggregating information from related operations, machines that can process these operations, and edges connecting the job to these machines. The initial representation \(\boldsymbol{h_{j_{i}}} \in \mathbb{R}^{d_{\mathcal{J}}}\) is updated by calculating attention coefficients that weigh the contributions of these connected nodes and edges. For instance, the attention coefficient \(\alpha_{mj_{ki}}\) between a job \(j_i\) and a machine \(m_k\) is computed as:
\begin{align}
&\alpha_{mj_{ki}}=\operatorname{softmax}  \Bigg( \nonumber  \\ 
& \frac{\left(\mathbf{W}_1^{\mathcal{JM}} \boldsymbol{h_{j_{i}}}\right)^{\top}\left(\mathbf{W}_2^{\mathcal{JM}} \boldsymbol{h_{m_{k}}} + \mathbf{W}_3^{\mathcal{JM}} \boldsymbol{h_{mj_{ki}}}\right )}{\sqrt{d_{\mathcal{JM}}^\prime}}\Bigg)
\end{align}
where \(\boldsymbol{W_1^{\mathcal{JM}}} \in \mathcal{R}^{d_{\mathcal{JM}}^\prime \times d_{\mathcal{J}}}\), \(\boldsymbol{W_2^{\mathcal{JM}}} \in \mathcal{R}^{d_{\mathcal{JM}}^\prime \times d_{\mathcal{M}}}\), and \(\boldsymbol{W_3^{\mathcal{JM}}} \in \mathcal{R}^{d_{\mathcal{JM}}^\prime \times d_{\mathcal{JM}}}\) are learned linear transformations, \(d_{\mathcal{JM}}^\prime\) is the dimension assigned to the hidden space of the embeddings, and $\boldsymbol{h_{mj_{ki}}}$ is the edge attribute between the job \(j_i\) and the machine \(m_k\). As mentioned in the previous section, $\log(\text{softmax}(a))$ is added to this edge for the critic. The attention coefficients \(\alpha_{jo_{ij}}\) between the job \(j_i\) and the operations \(o_{ij}\), and \(\alpha_{jj_{ii^\prime}}\) between the job \(j_i\) and other jobs \(j_{i^\prime}\), are computed in a similar manner.

Once the attention coefficients have been calculated, the updated embedding of \(j_i\), \(\boldsymbol{h_{j_{i}}^\prime}\), is computed as:
\begin{align}
\boldsymbol{h_{j_{i}}^\prime}= \mathbf{W}_1^{\mathcal{J}}\boldsymbol{h_{j_{i}}} + \sum_{j_{i^\prime} \in \mathcal{J} \; | \; i \neq i^\prime} \alpha_{jj_{ii^\prime}} \mathbf{W}_2^{\mathcal{J}} \boldsymbol{h_{j_{i^\prime}}} + \nonumber \\
\sum_{o_{ij} \in \mathcal{O}_{j_{i}}} \alpha_{jo_{ij}} \left(\mathbf{W}_4^{\mathcal{JO}} \boldsymbol{h_{o_{ij}}} + \mathbf{W}_5^{\mathcal{JO}} \boldsymbol{h_{jo_{ij}}}\right) + \nonumber \\
\sum_{m_k \in \mathcal{JM}_{j_{i}}} \alpha_{mj_{ki}} \left(\mathbf{W}_1^{\mathcal{JM}} \boldsymbol{h_{m_{k}}} + \mathbf{W}_2^{\mathcal{JM}} \boldsymbol{h_{mj_{ki}}}\right) 
\end{align}
where \(\mathcal{JM}_{j_i}\) is the set of edges connecting the job \(j_i\) with compatible machines, \(\boldsymbol{h_{m_{k}}}\) is the embedding of the machine \(m_k\), and \(\boldsymbol{h_{mj_{ki}}}\) are the features of the edge connecting the job and machine. We enhance the GNN's capability by utilizing multiple attention heads and stacking several layers, which enables the model to capture intricate node relationships.

Following the introduction of the general GNN, the architecture of the actor and critic is subsequently defined. The critic, which evaluates the actions taken by the policy, is defined by concatenating the mean of the job embeddings and the mean of the machine embeddings. This combined representation captures the overall state of both jobs and machines, allowing the critic to effectively assess the quality of the policy’s actions. The critic’s output is given by:

\begin{align}
Q(s_t', a_t)_{\phi} = \text{MLP}_{\phi}\left(\frac{1}{|\mathcal{J}|} \sum_{j_i \in \mathcal{J}} \boldsymbol{h_{j_i}}^\prime || \frac{1}{|\mathcal{M}|} \sum_{m_k \in \mathcal{M}} \boldsymbol{h_{m_k}}^\prime \right)
\end{align}
where $\text{MLP}_{\text{critic}}$ is a multi-layer perceptron, and $s_t'$ is the state to which the edge connecting the jobs and machines has been added $\log(\text{softmax}(a))$.  The embeddings calculated by the GNN are used to define the policy, represented as a probability distribution over the actions. Specifically, the probability of selecting an edge \(e \in \mathcal{JM}_t\) (connecting job \(j_i\) and machine \(m_k\)) is computed as:
\begin{align}
\mu_\theta(e | s_t) = \text{MLP}_{\theta}(\boldsymbol{h_{j_{i}}^L} || \boldsymbol{h_{m_{k}}^L} || \boldsymbol{h_{mj_{ki}}} )
\end{align}
\begin{align}
\pi_\theta(e | s_t) = \frac{\mu_\theta(e | s_t)}{\sum_{x \in \mathcal{JM}_t }{\mu_\theta(x | s_t)}}
\end{align}

\subsection{Instance Generation for Offline RL in Combinatorial Optimization Problems}

In the domain of combinatorial optimization, it is common practice to employ solvers or metaheuristic methods that can effectively identify optimal or near-optimal solutions for smaller problem instances \cite{blum2016construct}. Given the inherent difficulty in discovering optimal solutions through RL alone, these methods play a critical role in generating expert experiences that are subsequently leveraged as training data for offline RL.

A key contribution of our approach is the novel method we employ for generating and utilizing instances within the offline RL algorithm. A common challenge with algorithms that heavily depend on BC is their tendency to generalize poorly \cite{ross2010efficient}. This issue arises when the policy encounters states during testing that are underrepresented in the training data, leading to suboptimal performance. To overcome this limitation, we propose an instance generation strategy designed to create a more diverse and representative set of training experiences, thereby enhancing the policy’s ability to generalize to novel, unseen scenarios.

Our method comprises two key components. First, we start by defining an initial state configuration that sets the foundation for generating optimal solutions directly. From this predefined starting point, the policy can either follow a path leading to an optimal solution or encounter random actions that introduce variability. These random actions lead the policy through a broader range of potential states, some of which may deviate from the optimal path. 

\begin{figure}[H] \centering \includegraphics[width=0.8\linewidth]{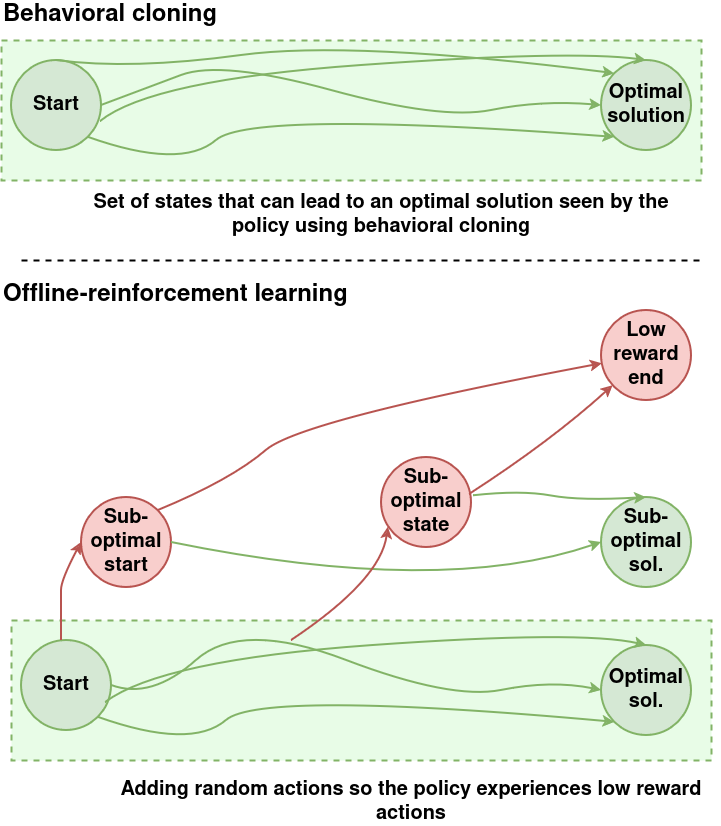} \caption{Comparison of BC, which focuses on optimal paths, with offline RL, which also explores suboptimal states and low-reward actions.} \label{fig:diagram_generation} \end{figure}

Second, to enhance the policy's ability to navigate suboptimal states and understand the consequences of low-reward actions, we integrate transitions that lead to such outcomes into the training dataset. These "negative" experiences are vital for teaching the policy how to identify and respond to suboptimal scenarios that may arise due to random actions or other disturbances. By being exposed to and learning from these low-reward scenarios, the policy becomes more adept at avoiding or mitigating such situations in the future, ultimately leading to improved overall performance. In Figure \ref{fig:diagram_generation} a diagram of the generation process is depicted. 

To effectively incorporate these diverse transitions into the learning process, we propose a modified loss function for the policy (actor) component of the offline RL algorithm. Specifically, the loss function is formulated as follows:
\begin{align}
\pi _{\theta} = \text{argmax}_{\pi} \mathbb{E}_{(s,a) \sim \mathcal{D}} & \Bigg[\lambda_{RL} Q(s, a)) - \\ \nonumber
& \delta(a)\lambda_{BC} \; \pi_{\mathcal{D}}(a | s) \log \left({\frac {\pi_{\mathcal{D}}(a | s)}{\pi(a | s)}}\right) \Bigg] 
\end{align}

where $\delta(a)$ is an indicator function that equals zero if the action is generated randomly and one if it is expert-generated. This configuration allows the policy to recognize and adapt to low-reward actions when they occur, while still emulating expert behavior.

Although generating these instances may require additional time due, the computational cost is limited to working with small instances. Furthermore, this approach can accelerate the training of the models, as in DRL, the algorithm typically needs to explore the action space to reach such solutions. By leveraging offline RL with pre-generated instances, we effectively reduce the time spent on exploration, enabling more efficient learning. The training process is outlined in Algorithm \ref{algo:TD3}. For clarity and to highlight the contribution of our algorithm, the more specific details of TD3 will not be shown. 

\begin{algorithm}[H]
\caption{Training models with H-ORL}
\label{algo:TD3}
\begin{algorithmic}[1]
\State \textbf{Input:} Initial policy parameters $\theta$, Q-function parameters $\phi_1$, $\phi_2$, dataset $\mathcal{D}$, and balancing hyperparameters $\lambda_{BC}$, $\lambda_{RL}$
\For{$epoch = 1$ to $n_{epochs}$}
    \For{each gradient step}
        \State Sample mini-batch of $N$ transitions $(s, a, r, s')$ from $\mathcal{D}$
        \State Construct augmented state $s''$ by adding $\log(\text{softmax}(\pi_{\mathcal{D}}(a | s)))$ as a job-machine edge attribute to $s$
        \State Compute target Q-value and update Q-functions by minimizing the squared error loss
        \If{delayed policy update (every $d$ steps)}
            \State Construct augmented state $s'''$ by adding $\log(\text{softmax}(\pi_{\theta}(a | s)))$ as an edge attribute to $s$
            \State Update policy by minimizing:
             \begin{align}
              \mathcal{L}_\pi(\theta) &= \frac{1}{N} \sum - -\Bigg[ -\lambda_{RL} Q(s, a)) + \\ \nonumber
& \delta(a)\lambda_{BC} \; \pi_{\mathcal{D}}(a | s) \log \left({\frac {\pi_{\mathcal{D}}(a | s)}{\pi(a | s)}}\right) \Bigg] 
\end{align}
        \EndIf
    \EndFor
\EndFor
\State \textbf{Output:} Optimized policy parameters $\theta$
\end{algorithmic}
\end{algorithm}


        

\section{Experiments}\label{sec:experimentalresult}
In this section, we present the experimental results to validate our method. First, we compare our offline RL method for heterogeneous graphs, which we refer to as H-ORL, with BC to demonstrate how incorporating reward information enhances the policy's performance. Next, we compare our algorithm for both the JSSP and FJSSP against various state-of-the-art DRL and BC methods using benchmark problems.

\subsection{Experimental setup}

\subsubsection{Configuration}

For our proposed method, we utilized Python 3.10 and PyTorch Geometric for implementing the graph neural networks \cite{Fey/Lenssen/2019}. For the constraint programming method, we chose OR-Tools\footnote{ We obtained the implementation from \url{https://github.com/google/or-tools/blob/stable/examples/python/flexible_job_shop_sat.py}.} due to its open-source availability and extensive use in scheduling problems \cite{da2019industrial}. Table \ref{tab:hyperparameters} details the configuration hyperparameters used for training our model, which are similar to other DRL-based approaches \cite{song2022flexible} \cite{ wang2023flexible}. The parameters used related to the TD3 algorithm are the default parameters as in the TD3 offline implementation from Offline RL-Kit \cite{offinerlkit}. The code will be published upon acceptance of the paper. 

\begin{table}[H]
\small
\centering
\caption{Model hyperparameters.}
\label{tab:hyperparameters}
\begin{tabularx}{\linewidth}{p{2cm} p{1cm} X}
\toprule
\textbf{Parameter} & \textbf{Value} & \textbf{Description} \\
\midrule
GNN layers & 5 & Number of hidden layers in the GNN \\
Attention heads & 3 & Number of attention heads in the GNN \\
MLP layers & 3 & Number of hidden layers in the MLP for actor and critic \\
Hidden dimension & 32 & Size of the hidden layers in GNN and MLPs \\
Learning rate & 0.0002 & Learning rate for the actor-critic \\
Optimizer & Adam & Optimizer used with default parameters \\
Epochs & 30 & Number of training epochs \\
Batch size & 128 & Size of each training batch \\
Training instances & 1000 & Number of instances used for training \\
\bottomrule
\end{tabularx}
\end{table}

For the environment, two parameters need to be set: first, the number of operations visible for each job (set to 10); and second, to limit the action space, we define \(t_e\) as the earliest time a machine can start a new operation. Actions are masked if their start time exceeds \(t_e \times p\), where \(p\) is set to 1.05.

\subsubsection{Dataset Generation and Benchmarks}

Two models were trained for the JSSP and FJSSP, with instances generated according to the specifications detailed in Table \ref{tab:datasetparams}.

\textbf{JSSP Dataset.} A set of 1000 instances was generated for training and 50 for validation, solved using OR-Tools with a ten-second limit. The generation followed the method proposed by Taillard \cite{taillard1993benchmarks}. The parameters used in generating the instances, such as the number of jobs, machines, and operations, were drawn from specific uniform distributions to ensure variability and train the model across different scenarios.

\textbf{FJSSP Dataset.} Similarly, 1000 instances were generated for training and 50 for validation, with each instance solved using OR-Tools within a one-minute time limit, following the method adapted from \cite{brandimarte1993routing}. The generation process involved sampling the number of jobs, machines, and operations per job from defined ranges, as well as varying the processing times to reflect different scenarios.

\begin{table}[H]
\small
\centering
\caption{Parameters employed to generate instances for the JSSP and FJSSP datasets.}
\label{tab:datasetparams}
\begin{tabularx}{\linewidth}{p{3cm} X X}
\toprule
\textbf{Parameter} & \textbf{JSSP} & \textbf{FJSSP} \\
\midrule
Number of jobs ($n_j$) & $U(10, 20)$ & $12$ \\
Number of machines ($n_m$) & $U(10, n_j)$ & $U(4, 9)$ \\
Operations per job & $U(n_m, n_m + 5)$ & $U(2, 9)$ \\
Operations per machine & $1$ & $U(2, m)$ \\
Mean processing time ($\overline{p}$) & N/A & $U(5, 10)$ \\
Processing time & $U(5, 90)$ & $\mathcal{U}(\overline{p}(1-d), \overline{p}(1+d))$ \\
Deviation ($d$) & N/A & 0.2 \\
\bottomrule
\end{tabularx}
\end{table}

\textbf{JSSP Test Benchmarks.} For the test set, we utilized the well-known JSSP benchmark dataset by Taillard \cite{taillard1993benchmarks}. This dataset includes 80 instances, ranging from 15 jobs and 15 machines (with 225 operations) to 100 jobs and 20 machines (with 2000 operations), providing a comprehensive assessment of the method's performance across different instance sizes.

\textbf{FJSSP Test Benchmarks.} For evaluating our FJSSP model, we used five benchmark datasets: Brandimarte \cite{brandimarte1993routing}, Hurink \cite{hurink1994tabu} (subdivided into three categories - vdata, edata, and rdata), and Dauzère-Pérès and Paulli \cite{dauzere1994solving}. These benchmarks are widely recognized and cover a range of instance sizes, from small to medium-large operations, providing a thorough assessment of the method's performance.

\subsubsection{Baselines and evaluation metric} The following outlines the baselines for each of the problems. We selected state-of-the-art DRL and BC methods as baselines to evaluate if our Offline RL approach offers a performance advantage.

\textbf{JSSP Baselines.} We compared our approach with several DRL-based state-of-the-art constructive methods, such as the method upon which this work is based, called MAS \cite{echeverria2024multi}, which only uses behavioral cloning and serves as a baseline for comparing the contribution of this paper; the method proposed in \cite{tassel2023end}, (referred to in this paper as RLCP); the algorithm introduced in \cite{ho2024residual} (ResSch); and the method proposed in \cite{yuan2023solving} (BiSch). Additionally, we compared with an improvement-based DRL methods proposed on \cite{zhang2024deep} called L2S. In all cases, the greedy strategy was used for comparison, which involves selecting the action with the highest probability assigned by the policy, except for \cite{zhang2024deep}, where the variant that improves a solution in 500 steps was used due to its lower computational cost and similar execution time.

\textbf{FJSSP Baselines.} Our method was evaluated against seven recent DRL-based state-of-the-art approaches: \cite{song2022flexible} (referred to as HGNN for proposing the use of heterogeneous graphs), \cite{wang2023flexible} (DANIEL), \cite{ho2024residual} (ResSch), \cite{yuan2024solving} (referred to as LMLP for proposing the use of a lightweight multi-layer perceptron), \cite{echeverria2025diverse} (EDSP), \cite{echeverria2025leveraging} (BC), and \cite{huang2024learning} (referred to as GGCT for its use of Graph Gated Channel Transformation). For \cite{song2022flexible}, \cite{echeverria2025diverse}, \cite{echeverria2025leveraging}, and \cite{wang2023flexible}, we utilized the models provided in their respective code repositories with the default parameters. All approaches used the proposed dataset benchmarks except for ResSch, GGCT, and LMLP, which did not use the Dauzère-Pérès and Paulli benchmark. A summary of the deep learning-based baselines is provided in Table \ref{tab:baselines} and a more detailed explanation of the baselines in Table \ref{tab:related}.

\begin{table}[H]
\small
\centering
\caption{Deep learning baselines used for comparison in JSSP and FJSSP datasets.}
\label{tab:baselines}
\begin{tabularx}{\linewidth}{p{2cm} p{2.5cm} p{1.5cm} X}
\toprule
\textbf{Dataset} & \textbf{Abbreviation} & \textbf{Year} &  \textbf{Reference} \\
\midrule
\multirow{6}{*}{JSSP} 
& RLCP & 2023 & \cite{tassel2023end} \\
& BiSch & 2023 &  \cite{yuan2023solving} \\
& ResSch & 2024 & \cite{ho2024residual} \\
& L2S & 2024 &\cite{zhang2024deep} \\
& MAS & 2024 & \cite{echeverria2024multi} \\
\midrule
\multirow{7}{*}{FJSSP} 
& HGNN & 2022 & \cite{song2022flexible} \\
& DANIEL & 2023 & \cite{wang2023flexible} \\
& EDSP &  2024 & \cite{echeverria2025diverse} \\
& ResSch & 2024 & \cite{ho2024residual} \\
& LMLP & 2024 & \cite{yuan2024solving} \\
& GGCT & 2024 & \cite{huang2024learning} \\
& BC &  2024 & \cite{echeverria2025leveraging} \\
\bottomrule
\end{tabularx}
\end{table}

\textbf{Meta-heuristic Baselines.} Additionally, comparisons with metaheuristic methods are included, though they are not directly comparable due to computational costs. Our approach is also evaluated against an enhanced metaheuristic \cite{falkner2022learning}, specifically the $NLS_{AN}$ policy, which has proven to outperform traditional JSSP metaheuristics on the Taillard benchmark. For the FJSSP, we have included metaheuristic algorithms like the Improved Jaya Algorithm (IJA) \cite{caldeira2019solving}, Two-Stage Genetic Algorithm (2SGA) \cite{rooyani2019efficient}, and Self-learning Genetic Algorithm (SLGA) \cite{chen2020self}.

The performance is assessed by calculating the optimal gap, with the makespan as the target for optimization. The optimal gap is defined as follows:
\begin{equation}
OG = \left( \frac{C_{\pi}}{C_{\text{ub}}} - 1 \right) \cdot 100
\end{equation}
In this equation, $C_{\pi}$ represents the makespan achieved by the policy, and $C_{\text{ub}}$ denotes the optimal makespan or the best-known makespan for the instance. While other evaluation criteria exist, such as minimizing total tardiness or maximizing throughput, minimizing the makespan remains the most widely used criterion in scheduling due to its general applicability and effectiveness in measuring efficiency.

\subsection{Experimental Results}

The experimental results are divided into two sections: 1) The first validates that including a reward-related term is beneficial compared to an approach exclusively based on BC. 2) The second presents results comparing the method with various public datasets and state-of-the-art methods.

\subsubsection{Offline Reinforcement Learning vs Behavioral Cloning}

\begin{figure*}[!t]
\centering
\includegraphics[width=\textwidth]{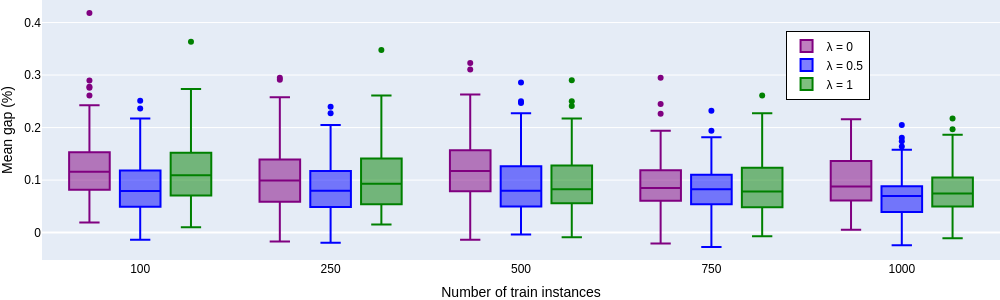}
\caption{Comparison using different numbers of training instances and lambda parameters.}
\label{fig:comparative_lambda}
\end{figure*}

First, we aim to answer whether using offline reinforcement learning benefits policy performance. As explained in \cite{kumar2022should}, generally, offline DRL is preferable to BC when dealing with random or highly suboptimal demonstrations, because the trained policy does not improve upon the demonstrations and only mimics them. However, this is not the case when optimal or expert demonstrations are available.

To experimentally validate that offline RL is indeed beneficial, policies have been generated based on different training set sizes (100, 250, 500, 750, and 1000), starting in a scenario where there are few instances. The goal of generating different policies with varying numbers of instances is to study how they evolve in different scenarios. For each training set, a policy has been generated with three different values of lambda $\lambda = { 0, 0.5, 1 }$, where the results on the validation set are shown in Figure \ref{fig:comparative_lambda}. Using a lambda value of zero is equivalent to using BC.

Several conclusions can be drawn from this figure. First, for every training set, the policy trained with $\lambda = 0.5$ yields the best median results. This suggests that including a term related to the reward is beneficial. Secondly, this term must be controlled and does not imply that a higher value is better. Indeed, with only 100 instances, the worst results are obtained with the policy trained with $\lambda = 1$. This may be due to the critic not having enough information to be trained properly, complicating policy generation.

Lastly, another interesting fact is that the policy based on BC does not improve as the number of instances increases, unlike the other policies. This may be due to the limitation of the network size in these types of problems. That is, in typical supervised learning problems, when it is assumed that the complexity of a neural network is not sufficient to capture the entire complexity of a dataset, a common practice is to increase the size of the network. However, in this case, this is not viable as an increase in network size would also increase execution time, making the solution no longer real-time. Therefore, strategies for generating the policy, such as the use of offline DRL, are necessary to achieve a better policy.

\subsection{Benchmarks Results}

\subsubsection{JSSP Results} This section presents an analysis of the performance of our method on the Taillard benchmark. Table VI summarizes the optimal gaps achieved by various methods, grouped into deep learning-based approaches and meta-heuristic methods. Each row in the table corresponds to a different problem size, defined by the number of jobs and machines. The columns represent the different methods, with the first five columns dedicated to deep learning-based methods and the last column representing an advanced meta-heuristic approach. In the appendix \ref{appendix:time}, a comparison of the execution times of the deep learning-based methods is made.

\begin{table*}[!t]
\scriptsize
\centering
\caption{Optimal gap comparison with an enhanced meta-heuristic, DRL methods, BC (MAS), and H-ORL in the Taillard benchmark.}
\label{tab:allbenchmarkstaillard}
\begin{tabular*}{\textwidth}{l @{\extracolsep{\fill}} r r r r r r r }
\toprule
\multirow{1}{*}{\textbf{Size}} & \multicolumn{6}{c}{\textbf{Deep learning methods}} & \multicolumn{1}{c}{\textbf{Meta-heuristic}} \\
\cmidrule(lr){2-7}
\cmidrule(lr){8-8}
& BiSch & ResSch & RLCP & MAS & L2S & \textbf{H-ORL} & $NLS_{AN}$ \\
\midrule
15$\times$15 & 17.85 & 13.70 & 16.27 & 13.54 & \textbf{9.30} & 14.76 & 10.32 \\
20$\times$15 & 20.59 & 18.00 & 19.75 & 13.96 & \textbf{11.60} & 13.98 & 13.18 \\
20$\times$20 & 19.53 & 16.50 & 18.61 & 13.40 & \textbf{12.40} & 14.09 & 12.95 \\
30$\times$15 & 22.39 & 17.30 & 18.29 & 14.84 & 14.70 & \textbf{13.13} & 14.91 \\
30$\times$20 & 23.32 & 18.10 & 22.81 & 17.40 & 17.50 & \textbf{17.18} & 17.78 \\
50$\times$15 & 16.03 & 8.40 & 10.13 & 7.08 & 11.00 & \textbf{5.38} & 11.87 \\
50$\times$20 & 17.41 & 11.40 & 14.03 & 9.51 & 13.00& \textbf{8.36} & 12.02 \\
100$\times$20 & 8.91 & 4.00 & 4.56 & 2.31 & 7.90 & \textbf{1.63} & 6.22 \\
Mean & 18.25 & 13.40 & 15.55 & 11.50 & 12.17 & \textbf{10.80} & 12.41 \\
\bottomrule
\end{tabular*}
\end{table*}

\begin{table*}[!t]
\tiny
\centering
\caption{Optimal gap comparison with meta-heuristic algorithms, DRL methods, BC, and H-ORL over five FJSSP public datasets.}
\label{tab:allbenchmarksfjssp}
\begin{tabular*}{\textwidth}{l @{\extracolsep{\fill}} r r r r r r r r r r r}
\toprule
\multirow{1}{*}{\textbf{Dataset}} & \multicolumn{8}{c}{\textbf{Deep learning methods}} & \multicolumn{3}{c}{\textbf{Meta-heuristic}} \\
\cmidrule(lr){2-9}
\cmidrule(lr){10-12}
 & HGNN & DANIEL & EDSP & LMLP & BC & GGCT & ResSch & \textbf{\textbf{H-ORL}} &IJA & 2SGA \footnotetext{This approach only uses the first 30 instances for the vdata benchmark.} & SLGA \\
\midrule
Brand. & 27.83 & 11.97 & 27.34 & 14.04 & 12.72 & 23.50 & 9.81 & \textbf{9.12} & 8.50 & 3.17 & 6.21 \\
Dauzere & 9.26 & 8.88 & 9.33  & 11.10 & 7.50 & - & - &\textbf{6.36} & 6.10 & - & - \\
edata & 15.53 & 13.73 & 17.09 & 15.54 & 12.02 & 15.28 &  13.20 & \textbf{11.49} & 3.90 & - & - \\
rdata & 11.15 & 12.17 & 12.78 & 12.14 & 8.26 &  10.18 & 9.60 & \textbf{8.02} & 2.70 & - & - \\
vdata & 4.25 & 5.40 & 3.38 & 5.35 & 2.84 & 3.82 &  3.80 & \textbf{2.61} & 4.60 & 0.39 & - \\
\bottomrule
\end{tabular*}
\end{table*}


An important finding from the table is that our method consistently achieves a lower average gap from the optimal solution compared to both DRL-based approaches and the enhanced metaheuristic $NLS_{AN}$ . This is especially evident in larger instances, particularly those with 30 jobs or more, where our method surpasses all others. Our approach outperforms other DRL methods in almost every instance group, except for the three smaller ones, where the L2S neural improvement method performs better.

We speculate that the superior performance of L2S and the metaheuristic method $NLS_{AN}$ in smaller instances, but their relatively poorer performance in larger ones, can be attributed to their reliance on state-space exploration. As the problem size increases, particularly in complex scheduling problems like JSSP, the state space expands exponentially, making it increasingly challenging for these methods to efficiently navigate and optimize within such a vast space. This limitation likely diminishes their effectiveness as the problem size scales up.

Moreover, it is particularly interesting that our method generalizes well to larger instances, despite being trained on smaller ones. This strong generalization ability indicates that our approach effectively captures the underlying structure of the problem, allowing it to maintain low optimal gap values even when applied to more complex and larger-scale scenarios.

\subsubsection{FJSSP Results}

Finally, we present the results for the FJSSP. Table \ref{tab:allbenchmarksfjssp} displays the average gaps achieved by various approaches across different benchmarks, with the values for the deep learning methods highlighted in bold.



The results drawn from the table indicate that our approach, H-ORL, consistently outperforms other DRL methods across the FJSSP benchmarks, similar to the results observed in the JSSP benchmark, where it demonstrated superior performance in reducing optimal gaps. While some meta-heuristic approaches still show better results on specific datasets, our method shows significant promise, especially in generating high-quality solutions. This underscores the potential of H-ORL not only for real-time solution generation but also as a strong candidate for producing superior solutions in more complex and varied problem instances.

\begin{figure*}[!t]
\centering
\includegraphics[width=\textwidth]{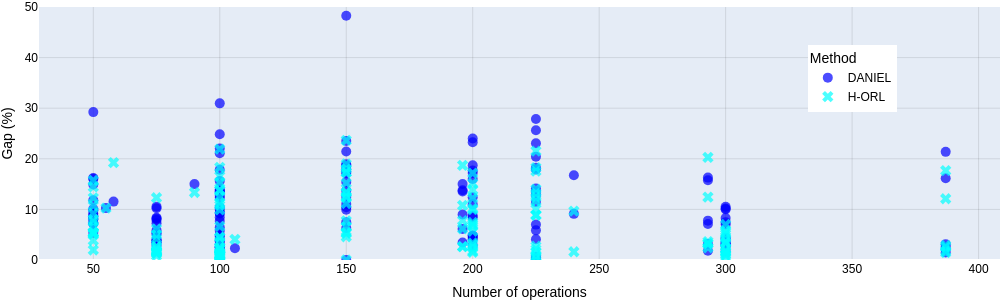}
\caption{Comparison of the performance gap between DANIEL and H-ORL across all instances of the FJSSP benchmarks.}
\label{fig:compare_wang}
\end{figure*}

Finally, we aim to compare the evolution of our method with respect to the size of the instances. In Figure \ref{fig:compare_wang}, a comparison is presented between the DANIEL method and the H-ORL method. The x-axis represents the number of operations for the instances, while the y-axis shows the resulting performance gap (\%) obtained by each method. Each dot corresponds to a specific instance, where blue circles denote results from the Wang method, and light blue 'x' markers represent the H-ORL method. As shown, the performance gaps vary with the number of operations, with certain groups of instances (for example, those around 100, 200, and 300 operations) showing more significant differences between the methods. This is consistent with the results obtained for the JSSP, where the best results were also achieved on larger instances.

\section{Conclusions and future work}\label{sec:conclusion}
This paper introduces a new offline RL action-value-based method tailored for combinatorial optimization problems, where the state is modeled as a heterogeneous graph and the action space is variable. By leveraging edge attributes to track actions at each step, our approach enhances the capability of current offline RL methods in handling complex problem structures. We also propose a novel loss function that effectively balances the goals of maximizing expected rewards and imitating expert solutions. This ensures that the policy is not only efficient but also grounded in proven strategies, leading to better generalization.

The efficacy of our method has been validated through extensive testing on five well-known benchmarks, including two for the JSSP and three for the FJSSP. Our approach consistently outperforms state-of-the-art methods, demonstrating its superiority in generating high-quality solutions across different problem settings. This work underscores the potential of offline RL in addressing challenging optimization problems with complex constraints. Future work could explore other offline RL methods to further enhance performance or to apply this approach to other combinatorial optimization problems. In particular, we will investigate more suitable algorithms for discrete action spaces, such as BCQ \cite{fujimoto2019off}, which could lead to improvements in handling discrete state-action relationships more effectively.

\section*{Acknowledgements}
This work was partially financed by the Basque Government through their Elkartek program (SONETO project, ref. KK-2023/00038) and the Gipuzkoa Provincial Council through their Gipuzkoa network of Science, Technology and Innovation program (KATEAN  project, ref. 2023-CIEN-000053-01). R. Santana acknowledges partial support by the Research Groups 2022-2024 (IT1504-22) and the Elkartek Program (KK-2023/00012, KK-2024/00030) from the Basque Government, and the PID2022-137442NB-I00 and PID2023-149195NB-I00 research projects from the Spanish Ministry of Science.

\bibliographystyle{unsrt}
\bibliography{main}

\appendix
\section{Node and edge features}\label{appendix:nodefatures}
In this appendix, we detail the features of the nodes and edges in the state representation.

For job-type and operation-type nodes, the features are:
\begin{table*}[!t]
\scriptsize
\centering
\caption{Average computation time for the Taillard benchmark.}
\label{tab:times}
\begin{tabular*}{\textwidth}{l @{\extracolsep{\fill}} r r r r r r r r r}
\toprule
Method & $15\times15$ & $20\times15$ & $20\times20$ & $30\times15$ & $30\times20$ & $50\times15$ & $50\times20$ & $100\times20$ & Mean \\
\midrule
BiSch & 0.78 & 1.04 & 1.39 & 1.62 & 2.49 &  3.68 & 6.41 & 30.35 & 5.97\\
ResSch & 0.50 & 0.89 & 0.97 &  19.93 & 2.27 & 4.75 & 5.93 & 19.76 & 4.63 \\
RLCP & 5.60 & 6.32 & 7.29 & 8.99 & 10.90 & 16.96 & 20.64 & 63.75 & 18.81\\
MAS & 3.10 &  3.78 & 5.02 & 5.41 & 6.72 & 12.07 & 13.83 & 58.24 & 13.52 \\
L2S & 9.30 & 10.10 & 10.90 & 12.70& 14.00 & 16.20 & 22.80 & 50.20 & 18.28\\
H-ORL & 3.25 & 3.95 & 5.28 & 5.67 & 6.95 & 12.45 & 14.22 & 60.87 & 10.08\\
\bottomrule
\end{tabular*}
\end{table*}

\begin{itemize}
\item For jobs a binary indicator $b_j \in \{0, 1\}$ that indicates if the job is completed and for operations a binary indicator $b_o \in \{0, 1\}$ that indicates if the operation is ready.
\item For jobs and operations, the completion time of the last operation: Tracks job progress and aids in scheduling.
\item For jobs and operations, the number of remaining operations: Indicates the remaining workload.
\item For jobs and operations, sum of average processing times of remaining operations: Estimates the total remaining workload by adding up the average processing times of all unfinished operations.
\item For jobs, the um of average processing times of remaining operations: Assesses remaining workload.
\item For jobs, the mean processing time: Estimates operation duration.
\item For jobs, the minimum processing time: Highlights the quickest possible execution time.
\item For jobs, the ratio of mean processing time to sum of remaining operations' average processing times: Aids in prioritization.
\end{itemize}

For machine-type nodes, the features are:

\begin{itemize}
\item Total number of assignable operations to machine $m$: $|\mathcal{O}_m|$, the total number of operations that can be assigned to machine $m$.
\item Total number of operations that can only be assigned to machine $m$: $|\mathcal{O}_m^{\text{unique}}|$, the number of operations that can only be assigned to machine $m$.
\item Number of operations that can be immediately assigned to machine $m$: $|\mathcal{O}_m^{\text{immediate}}|$, the number of operations that can be immediately assigned to machine $m$.
\item Machine final completion time:$t_{\text{final}}^m$, the completion time of the last operation assigned to machine $m$.
\item Minimum machine option time: $\min(\mathcal{O}_m^{\text{options}})$, the minimum processing time of the available operations for machine $m$.
\item Mean machine option time: $\frac{1}{|\mathcal{O}_m^{\text{options}}|} \sum_{o \in \mathcal{O}_m^{\text{options}}} p_{o,m}$, the average processing time of the available operations for machine $m$.
\item Free machine time: $t_{\text{final}}^m - t_{\text{min}}$, the free time after the machine completes its last operation.
\item Machine availability flag: Binary feature $f_m = 0$ if the machine is unavailable, otherwise $f_m = 1$.
\item Machine utilization rate: $\frac{\text{occupations}^m}{t_{\text{final}}^m}$, the occupation rate of machine $m$.
\end{itemize}



Edges in the graph are characterized as follows:

\begin{itemize}
\item Undirected edges between machines and operations: Represent machine-operation compatibility and carry features like processing time.
\item Directed edges from operations to jobs: Indicate job-operation relationships.
\item Directed edges between operations: Represent precedence constraints.
\item Directed edges from machines to jobs: Connect machines to jobs' first pending operations and carry processing time features.
\item Connections between jobs and machines: Represent interdependence and mutual influence.
\end{itemize}

Only two edge types carry specific features: operation-machine and job-machine edges. Features for operation-machine edges include:

\begin{itemize}
\item Processing time $p_{o,m}$ for operation $o$ on machine $m$.
\item Normalized processing time $\frac{p_{o,m}}{|\mathcal{M}_o|}$, where $|\mathcal{M}_o|$ is the number of capable machines.
\item Normalized machine processing capability $\frac{p_{o,m}}{|\mathcal{O}_m|}$, where $|\mathcal{O}_m|$ is the number of operations machine $m$ can process.
\item Processing time divided by the sum of the average of the processing times of remaining operations.
\end{itemize}

For job-machine edges, features are analogous, focusing on the delay or gap caused by machine waiting times between operations, leading to idle time.

\section{Computation times comparison}\label{appendix:time}

Table \ref{tab:times} and \ref{tab:times_fjssp} present the computation times for deep learning methods applied to the Taillard and FJSSP benchmark problems. For the JSSP, execution times were obtained from the respective papers for all methods, except for RLCP, where its open-source implementation was used since it did not provide detailed execution times for different instance sizes. Most methods exhibit similar execution times. However, it is important to note that a precise comparison of execution times largely depends on the implementation and hardware used.

For the FJSSP, open-source implementations of the methods were used (except from ResSch), but we were unable to obtain times for LMLP or GGCT as they did not publish inference times or provide their implementations. In this case, there are no major differences between the methods since all of them utilize similar types of neural networks and model the problem in a comparable way.

\begin{table}[H]
\small
\centering
\caption{Average computation time for the FJSSP benchmarks.}
\label{tab:times_fjssp}
\begin{tabular}{l @{\extracolsep{\fill}} r r r r r r}
\toprule
Method & Brandimarte & Dauzere & edata & rdata & vdata \\
\midrule
HGNN  & 1.56 & 3.00 & 1.67 & 1.67 & 1.61 \\
DANIEL &  1.23 & 2.63 & 1.37 & 1.35 & 1.34\\
EDSP &  1.33 & 2.29 & 1.23 & 1.26 & 1.62 \\
BC & 1.63 & 4.35 & 1.78 & 1.81 & 2.14\\
ResSch & 0.85 & - & 0.80  & 0.63 & 0.98 \\
H-ORL & 1.93 & 3.93 & 1.86 & 1.96 & 2.21 \\
\bottomrule
\end{tabular}
\end{table}

\end{multicols*}

\end{document}